# Phase of Flight Classification in Aviation Safety using LSTM, GRU, and BiLSTM: A Case Study with ASN Dataset

Aziida Nanyonga
School of Engineering and Information Technology
University of New South Wales
Canberra, Australia
a.nanyonga@adfa.edu.au

Hassan Wasswa
School of Engineering and Information Technology
University of New South Wales
Canberra, Australia
h.wasswa@adfa.edu.au

Graham Wild
School of Engineering and Information Technology
University of New South Wales
Canberra, Australia
g.wild@adfa.edu.au

*Abstract*—Safety is the main concern in the aviation industry, where even minor operational issues can lead to serious consequences. This study addresses the need for comprehensive aviation accident analysis by leveraging natural language processing (NLP) and advanced AI models to classify the phase of flight from unstructured aviation accident analysis narratives. The research aims to determine whether the phase of flight can be inferred from narratives of post-accident events using NLP techniques. The classification performance of various deep learning models was evaluated. For single RNN-based models, LSTM achieved an accuracy of 63%, precision 60%, and recall 61%. BiLSTM recorded an accuracy of 64%, precision 63%, and a recall of 64%. GRU exhibited balanced performance with an accuracy and recall of 60% and a precision of 63%. Joint RNN-based models further enhanced predictive capabilities. GRU-LSTM, LSTM-BiLSTM, and GRU-BiLSTM demonstrated accuracy rates of 62%, 67%, and 60%, respectively, showcasing the benefits of combining these architectures. To provide a comprehensive overview of model performance, single and combined models were compared in terms of the various metrics. These results underscore the models' capacity to classify the phase of flight from raw text narratives, equipping aviation industry stakeholders with valuable insights for proactive decision-making. Therefore, this research signifies a substantial advancement in the application of NLP and deep learning models to enhance aviation safety.

## I. INTRODUCTION

Aviation safety is of paramount importance in the modern era of air travel [1], [2]. Accurate and timely decision-making is vital to ensure the safety and reliability of air transportation systems [3]. One critical aspect of aviation safety is the precise classification of an aircraft's phase of flight, which encompasses taxiing, take-off, cruising, descending, and landing [4]. This classification enables informed and proactive decision-making, ultimately contributing to the prevention of accidents and incidents.

The significance of phase of flight classification lies in its ability to provide a real-time understanding of an aircraft's operational state [5]. For instance, during take-off and landing phases, aircraft are particularly susceptible to various risks, including runway incursions, wind shear, and bird strikes. Timely identification of these phases allows air traffic controllers to implement appropriate safety measures and ensures that pilots receive the relevant information to make informed decisions [6], [7].

The Aviation Safety Network (ASN) has been a cornerstone resource for collecting, documenting, and disseminating aviation safety data from across the globe. The ASN dataset is a comprehensive set of transport category aircraft accidents. This data presents an opportunity to conduct research and analysis to enhance aviation safety measures [8].

In recent years, the field of deep learning, particularly recurrent neural networks (RNNs), has made significant strides in handling sequential data and time-series analysis [9], [10]. Long Short-Term Memory (LSTM) network [9], Gated Recurrent Unit (GRU) networks [11], and Bidirectional LSTM (BiLSTM) networks [12] have emerged as powerful tools for various sequential data classification tasks.

The motivation behind this research is two-fold. First, it arises from the pressing need to leverage cutting-edge machine learning techniques to enhance aviation safety protocols. Accurate and automated classification of an aircraft's phase of flight can significantly contribute to minimizing the risk of accidents and improving the overall safety of air travel. Second, the motivation stems from the potential of deep learning techniques to address complex and dynamic challenges in aviation safety [13]. While traditional rule-based methods have been employed for phase-of-flight classification, they may lack the adaptability and scalability needed to handle the increasing volume and complexity of aviation data. Deep learning models, with their ability to capture intricate patterns in data, offer a promising alternative [14].

The primary objective of this study is to assess the efficacy of NLP and Deep Learning techniques in categorizing flight phases within safety occurrence reports sourced from the ASN database. To fulfill this objective, we harnessed advanced deep learning architectures, including LSTM, BiLSTM, and GRU models, and combinations thereof, as suggested by [15]. These models underwent extensive training to deduce flight phase information from the unstructured text narratives present in the safety occurrence reports. To gauge model performance, we employed evaluation metrics encompassing accuracy, precision, recall, and F1-score. Our overarching goal is to showcase that NLP and Deep Learning models possess the capability to reliably infer flight phase information from raw text narratives, thereby establishing a robust foundation for more comprehensive safety occurrence analysis.

The structure of this paper is as follows: Section II provides a review of the existing literature on flight phase classification in aviation safety research, Section III gives an account of the methodology employed in this study, including data preprocessing, model selection, training, and evaluation, Section IV presents the results of our experiments showcasing the performance of the deep learning models in flight phase classification and Finally, Section V concludes the paper by summarizing the key findings and their significance in enhancing aviation safety.

## II. RELATED WORK.

Research in the field of aviation safety and phase of flight classification has seen notable developments, with an increasing focus on leveraging NLP techniques for accurate and automated classification. In this section, we review relevant studies and categorize them based on their contributions to this domain.

Early efforts in the phase of flight classification predominantly relied on rule-based approaches [16], [17]. These methods involved the formulation of specific rules and heuristics based on expert knowledge and domain-specific information. While these approaches were effective to some extent, they often struggled to adapt to evolving aviation environments and handle the complexity of unstructured text narratives from the Aviation Safety Network (ASN) dataset.

In recent years, machine learning techniques have gained prominence for phase-of-flight classification tasks [4], [7], [18]. These studies have demonstrated the potential of supervised learning algorithms, such as Support Vector Machines (SVM) and Random Forests, in automatically categorizing phases of flight based on textual descriptions. While machine learning models exhibited improved performance overrule-based methods, they faced challenges in capturing nuanced patterns and context from text narratives.

With the advent of deep learning and NLP, there has been a shift towards more sophisticated methods for phase-of-flight classification. Recurrent Neural Networks (RNNs) have shown promise in processing sequential textual data [19]. Specifically, Long Short-Term Memory (LSTM) networks and Gated Recurrent Unit (GRU) networks have been applied to extract temporal dependencies and semantics from ASN narratives [8]. Bidirectional LSTM (BiLSTM) architectures, which consider both past and future context, have further improved the accuracy of the phase of flight classification [20]. These models have the ability to capture complex dependencies in text data and learn contextual information that can be crucial for distinguishing between phases of flight.

Some recent studies have explored the combination of textual data with other modalities, such as flight data and audio transcripts, to improve the phase of flight classification [21]. These multimodal approaches aim to harness complementary information sources to enhance the accuracy and robustness of classification models.

In their study, Nanyonga et al [7] investigated the classification of safety occurrences in air transport using NLP and artificial intelligence (AI) models. The study utilized ResNet and sRNN deep learning models to classify flight phases based on unstructured text narratives of safety occurrence reports from the National Transportation Safety Board (NTSB). The research found that both models achieved an accuracy exceeding 68%, surpassing the random guess rate for the seven-class classification problem. Notably, the sRNN model outperformed the simplified ResNet architecture, suggesting the effectiveness of NLP and deep learning models in extracting flight phase information from raw text narratives.

Puranik et al [22] highlighted the application of supervised machine learning techniques in aviation safety, particularly for the prediction of safety-critical landing metrics. The study addressed the challenges of real-time risk identification by developing an offline-online framework using flight data from the approach phase. The framework leveraged Random Forest regression to predict landing true airspeed and ground speed, achieving robust and fast predictions suitable for online applications.

Fala et al [4] addressed the accurate identification of phases of flight in general aviation. The study explored dimensionality reduction algorithms and clustering techniques for phase identification. The research found that combinations of dimensionality reduction methods did not significantly impact phase identification and that both K-means and Gaussian mixture models were suitable for identifying flight phases in general aviation flights.

Our study extends the existing work in several ways. We explore the use of advanced Deep Learning models in flight phase classification, including LSTM, BiLSTM, and GRU, and a combination of these models specifically on ASN datasets that have not been extensively studied in this context.

## III. METHODOLOGY

In this section, we outline the methodology employed in this research to classify flight phases within safety occurrence reports from the ASN using NLP and Deep Learning techniques. To implement the proposed approach, our research encompassed several distinct phases, including data acquisition, text processing, and classification, as illustrated in Figure 1.

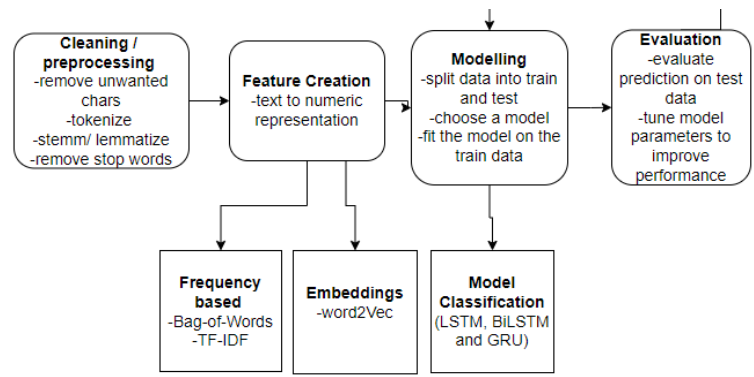

Fig. 1. Methodological framework

### A. Data Collection

Aviation incident and accident investigation reports are regularly compiled and disseminated by various organizations, such as the Australian Transport Safety Bureau (ATSB), the National Transportation Safety Board (NTSB), and the Aviation Safety Network (ASN). In this study, we focused on utilizing aviation incident and accident investigation reports from the ASN website spanning from 01/01/2000 to 12/31/2020 which is publicly available from the source: https://aviation-safety.net/. Specifically, our analysis concentrated on incidents where investigations had been completed, resulting in a dataset comprising 4,372 records following data preprocessing and cleaning. From each report, we extracted the 'Narrative' and ' phase of flight'

(POF) fields, which were used for training and validating our deep learning models.

*B. Text Pre-processing*

Machine learning models are not inherently designed to comprehend raw text data, necessitating the transformation of human-readable text into a numerical format that can be interpreted by the models. To facilitate this transformation, we employed the Keras deep learning library, which offers a comprehensive suite of deep learning models and various model layers. Additionally, Keras provides advanced modules for text preprocessing, including the Tokenizer module, which generates tokens and sequence vectors for input text. To encode categorical data, such as phase (s) of flight labels (e.g., take-off, initial climb, approach, landing, maneuvering, cruise, descent, standing, taxiing, and others), we employed the categorical module in Keras which uses one-hot to map categorical entries to numerical values encoding. Special characters, punctuation, stop-words, and word lemmatization were handled using the spacy library, specifically designed for text processing tasks, including named entity recognition and word tagging. Spacy incorporates a comprehensive list of special characters, punctuation marks, and stop-words and is regularly updated to accommodate evolving requirements. Leveraging these tools, we processed each input text narrative, transforming it into a representative sequence/vector with a fixed length of 2000. For narratives with a word count less than 2000, numeric sequences were padded with zeros, while those exceeding 2000 were truncated. The corpus vocabulary was limited to 100,000 terms. To split the dataset into train, and test subsets, we utilized Scikit-learn's train-test-split module. All experiments were conducted using Python as the programming language and Jupyter Notebook as the code editor.

*C. Text Classification*

The processed dataset was randomly partitioned into 80% train and 20% test sets. Additionally, 10% of the train set was held out for model validation during each epoch of training. Various deep learning models, including LSTM, BLSTM, GRU, and combinations of (LSTM+GRU, BLSTM+GRU, LSTM+BLSTM, and LSTM+BiLSTM+ GRU), were trained and evaluated, with their performance compared against other models. The Adam optimizer was employed for model optimization.

*D. Deep Learning Model Architecture*

To maintain consistency across models, except for minor variations in combined models, a standardized architecture was adopted. This architecture consisted of an embedding layer, hidden layers, and an output layer. Rectified Linear Unit (ReLU) activation functions were employed for all hidden layers, while the output layer utilized SoftMax as the activation function. The predicted class was generated using the argmax function, which identifies the index corresponding to the entry with the highest probability in the SoftMax output. Figure 2 illustrates the general deep learning architectures employed in this study for a single RNN-based deep learning algorithm (a) and when two distinct RNN-based deep learning algorithms are combined (b).

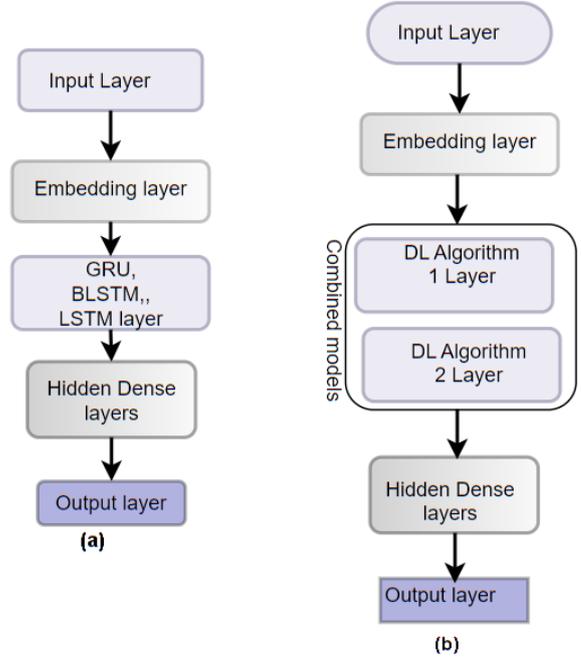

Fig. 2. Deep learning architectures: (a) single DL algorithm and (b) using two different DL Algorithm

This consistency of model architectures provided a solid foundation for training and evaluation of the deep learning models, enabling a fair comparison of their performance and the accurate classification of flight phases based on unstructured safety occurrence narratives.

*E. Model Performance Evaluation*

This section outlines the assessment criteria adopted for evaluating the models' performance in this study. Our primary focus is on multi-class classification, with performance evaluation centered on the accuracy of predictions across multiple classes. To ensure a comprehensive evaluation, we applied a set of well-established prediction performance metrics, including precision, recall, F1-score, and accuracy, as referenced in prior literature [23], [24]. These metrics, elaborated upon in Table 1, offer a thorough evaluation of the models' classification performance, with TP representing true positives, TN signifying true negatives, FN denoting false negatives, and FP indicating false positives.

TABLE I. PERFORMANCE EVALUATION METRICS

| Metrics | Formula | Evaluation focus |
|---|---|---|
| Precision (p) | $\frac{TP}{TP+FP}$ | Correctly predicted positives in a positive class |
| Recall (r) | $\frac{TP}{TP+TN}$ | Fraction of positive patterns correctly classified |
| F1-score (F) | $\frac{2*precision*recaal}{precision+recall}$ | Weighted average score of precision and recall |
| Accuracy (acc) | $\frac{TP+TN}{TP+FP+TN+FN}$ | Total number of instances predicted correctly |

## IV. RESULTS AND DISCUSSION

The experiments conducted in this study have showcased the remarkable capacity of NLP and advanced deep learning models to extract essential information from unstructured textual narratives of pre- and post-accident events, enabling the accurate classification of flight phases within safety occurrence reports. Our evaluation encompassed a diverse array of deep learning architectures, including GRU, LSTM, and BiLSTM, and joint RNN architecture, where the model incorporated multiple RNN layers of different architectures (e.g., GRU with LSTM, as GRU+LSTM, LSTM with BiLSTM as LSTM+BiLSTM and GRU with LSTM and BiLSTM as LSTM+BiLSTM+GRU). These models demonstrated exceptional power in learning intricate patterns from text narratives, which would lead to significant advancement in aviation safety analysis.

### A. Performance of Single RNN-Based Deep Learning Models

This section delves into the individual performance metrics of each model, comprising training accuracy, validation accuracy, testing accuracy, precision, recall, and F1 measure. It is worth noting that, in this context, a "single RNN-based deep learning model" refers to a model containing just one RNN-based layer (as depicted in Fig. 3). Conversely, a "joint RNN-based deep learning model" pertains to a model that incorporates multiple RNN-based layers of different architectures (as visualized in Fig. 4). Table II provides a record of the performance achieved by each of the three single RNN-based deep learning models employed in this study, are further visualized in Figure 3.

TABLE II. SINGLE RNN-BASED MODEL PERFORMANCE

| Modals | Precision (%) | Recall (%) | F1-Score (%) | Accuracy (%) |
|---|---|---|---|---|
| LSTM | 0.60 | 0.61 | 0.59 | 0.63 |
| BiLSTM | **0.63** | **0.64** | **0.63** | **0.64** |
| GRU | 0.63 | 0.60 | 0.61 | 0.60 |

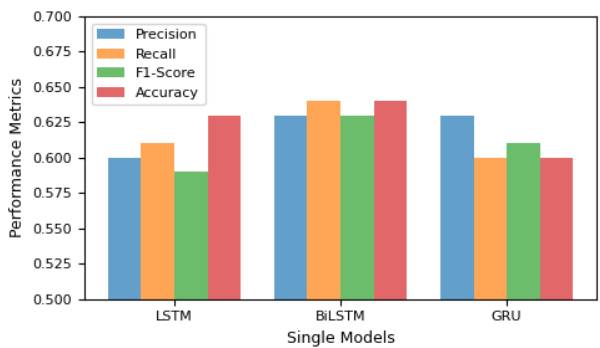

Fig. 3. Classifiaction performance of single models

### B. Performance of Joint RNN-Based Models

In addition to the single RNN-based deep learning models, we evaluated joint RNN-based models using the same dataset. Table III presents the prediction results from four different combinations of the three RNN-based models assessed in this study. The performance results of these joint models are visually presented in Figure 4.

TABLE III. JOINT RNN MODEL PERFORMANCE

| Modals | Precision (%) | Recall (%) | F1-Score (%) | Accuracy (%) |
|---|---|---|---|---|
| GRU-LSTM | 0.61 | 0.62 | 0.60 | 0.62 |
| GRU+BiLSTM+LSTM | 0.67 | 0.60 | 0.60 | 0.60 |
| LSTM+BiLSTM | **0.66** | **0.67** | **0.66** | **0.67** |
| GRU+BiLSTM | 0.70 | 0.62 | 0.64 | 0.62 |

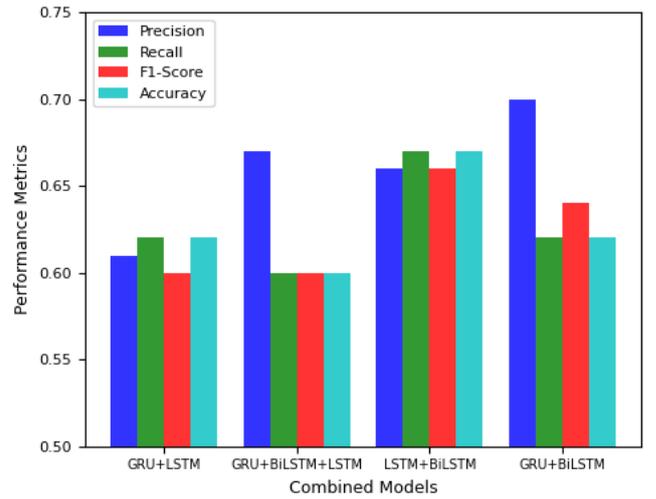

Fig. 4. Classification performance from combined models

These results underscore the potency of NLP and Deep Learning models in managing the complexity of aviation safety reports and extracting vital flight phase information. Notably, the LSTM+BiLSTM model exhibited the highest accuracy and precision, signifying their suitability for this task. Visualizations in Fig. 5 and Fig. 6 offer insights into the training behavior and performance of different deep learning models from both single and combined architectures, in regard to validation accuracy and validation loss, over various training epochs.

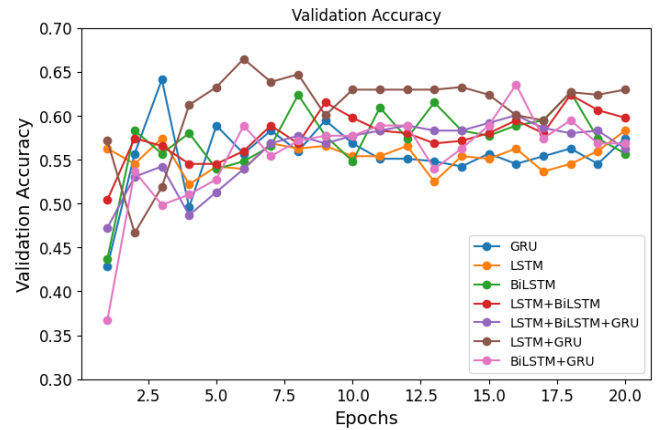

Fig. 5. Validation accuracy performance for both single and combined models

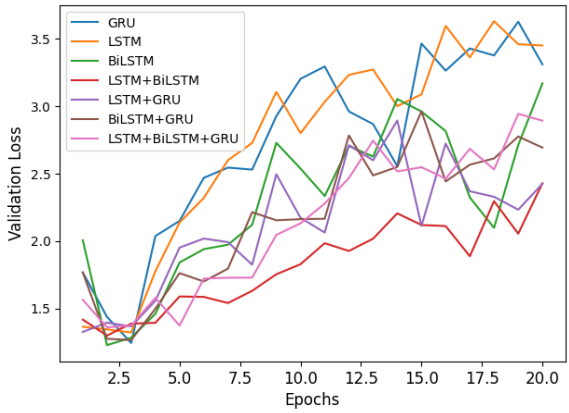

Fig. 6. Validation Loss for both single and combined models

```
Classification Report
              precision    recall  f1-score   support

    Approach       0.45      0.47      0.46        89
     Enroute       0.59      0.47      0.52       149
     Landing       0.82      0.85      0.84       338
    Standing       0.65      0.64      0.65        81
     Takeoff       0.53      0.70      0.60        87
        Taxi       0.46      0.41      0.44        41
     Unknown       0.24      0.21      0.22        72

    accuracy                           0.67
   macro avg       0.56      0.57      0.56       122
weighted avg       0.66      0.67      0.66       122
```

Fig. 7. Classification report for the best model LSTM+BiLSTM

Figure 7 provides an extract depicting the classification report, encompassing accuracy, precision, recall, and F1-score, for the best-performing model, LSTM+BiLSTM. This extract also elucidates the distribution of test instances among distinct phases of flight (POF) entries, as evident in the support column.

our research showcases the potential of deep learning models in effectively classifying flight phases based on unstructured text narratives. Joint models that amalgamate different RNN architectures show promise in enhancing classification accuracy. The model selection should be tailored to the specific application's requirements, considering the trade-offs between precision and recall.

## V. CONCLUSION

In an era where aviation safety is paramount, the demand for automated and precise safety occurrence analysis is escalating. This study has unequivocally unveiled the remarkable potential of NLP and Deep Learning methodologies in revolutionizing the classification of flight phases within safety occurrence reports. Leveraging a dataset comprising 4,327 reports sourced from ASN, we harnessed the prowess of advanced Deep Learning architectures, including LSTM, Bidirectional LSTM (BiLSTM), and GRU. These models exhibited remarkable efficiency in extracting crucial flight phase information from the intricate realm of unstructured textual narratives, marking a substantial leap forward in aviation safety analysis.

While our models achieved commendable performance in the context of seven-class classification, surpassing the 14% random guess threshold, it is essential to acknowledge the potential for further enhancement. The limited dataset size undoubtedly played a role, and future research endeavours should encompass the utilization of more extensive datasets, including the amalgamation of data from various sources such as the NTSB and ATSB. This approach holds the promise of refining model performance and elucidating the superiority of hybrid models in phase of flight classification. Notably, BiLSTM outperformed all single models, while the combination of LSTM and BiLSTM showcased superior performance among the joint models. This study underscores the transformative potential of NLP and Deep Learning in advancing aviation safety analysis, with ample opportunities for further exploration and refinement in the quest for safer skies.